\documentclass[11pt,a4paper]{article}
\usepackage[hyperref]{acl2019}
\usepackage{times}
\usepackage{latexsym}
\usepackage{url}
\usepackage{amsmath}
\usepackage{amsthm}
\usepackage{amsfonts}
\usepackage{amssymb}
\usepackage{graphicx}
\usepackage{booktabs}
\usepackage{placeins}
\usepackage{multirow}
\usepackage{tabularx}
\usepackage[ruled, linesnumbered]{algorithm2e}
\usepackage{epstopdf}
\usepackage{color}
\usepackage{subcaption}
\usepackage[T1]{fontenc}
\usepackage{amsmath,amsfonts,bm}
\usepackage{balance}
\usepackage{enumitem}
\usepackage{xcolor}
\usepackage{balance}
\usepackage{wrapfig}
\usepackage{subcaption}
\usepackage{enumitem}
\usepackage{graphics}
\usepackage{url}
\usepackage{makecell}
\usepackage[utf8]{inputenc}
\usepackage[english]{babel}

\newtheorem{theorem}{Theorem}

\usepackage{soul}

\aclfinalcopy 

\title{Incorporating Syntactic and Semantic Information in \\ Word Embeddings using Graph Convolutional Networks}

\author{Shikhar Vashishth$^1$ \quad Manik Bhandari$^{2*}$  \quad Prateek Yadav$^3$\thanks{ \ \ Contributed equally to the work.} \quad \\ \textbf{Piyush Rai}$^4$ \quad  \textbf{Chiranjib Bhattacharyya}$^1$ \quad 	\textbf{Partha Talukdar}$^1$\\
	\vspace{-4 mm} \\
	$^1$Indian Institute of Science, $^2$Carnegie Mellon University\\
	$^4$Microsoft Research, $^4$IIT Kanpur \\
	{\tt \small \{shikhar,chiru,ppt\}@iisc.ac.in}, {\tt \small mbhandar@andrew.cmu.edu} \\
	\vspace{-5.5 mm} \\
	{\tt \small t-pryad@microsoft.com, piyush@cse.iitk.ac.in} \\
}

\date{}

\begin{document}
\maketitle

 \newcommand{\refalg}[1]{Algorithm \ref{#1}}
\newcommand{\refeqn}[1]{Equation \ref{#1}}
\newcommand{\reffig}[1]{Figure \ref{#1}}
\newcommand{\reftbl}[1]{Table \ref{#1}}
\newcommand{\refsec}[1]{Section \ref{#1}}

\newcommand{\reminder}[1]{\textcolor{red}{[[ #1 ]]}\typeout{#1}}
\newcommand{\reminderR}[1]{\textcolor{gray}{[[ #1 ]]}\typeout{#1}}

\newcommand{\add}[1]{\textcolor{red}{#1}\typeout{#1}}
\newcommand{\remove}[1]{\sout{#1}\typeout{#1}}

\newcommand{\m}[1]{\mathcal{#1}}
\newcommand{\method}{SynGCN}
\newcommand{\methods}{WG}
\newcommand{\methodside}{SemGCN}
\newcommand{\methodsyn}{SynGCN}
\newcommand{\methodsidefull}{Semantic-GCN}
\newcommand{\methodsynfull}{Syntactic-GCN}

\newcommand{\problem}{DD}
\newcommand{\problemfull}{Document Dating}

\newcommand{\tensor}{\mathcal{X}}
\newcommand{\Real}{\mathbb{R}}

\newcommand{\tuples}{\mathbb{T}}

\newcommand{\argmax}{arg\,max}

\newcommand\norm[1]{\left\lVert#1\right\rVert}

\newcommand{\note}[1]{\textcolor{blue}{#1}}

\newcommand*{\Scale}[2][4]{\scalebox{#1}{$#2$}}%
\newcommand*{\Resize}[2]{\resizebox{#1}{!}{$#2$}}%
\definecolor{officegreen}{rgb}{0.0, 0.5, 0.0}
\def\mat#1{\mbox{\bf #1}}

\begin{abstract}
Word  embeddings have been widely adopted across several NLP applications. Most existing word embedding methods utilize \emph{sequential context} of a word to learn its embedding. While there have been some attempts at utilizing \emph{syntactic context} of a word, such methods result in an explosion of the vocabulary size. In this paper, we overcome this problem by proposing \method{}, a flexible Graph Convolution  based method for learning word embeddings. \method{} utilizes the dependency context of a word without increasing the vocabulary size.
Word embeddings learned by \method{} outperform existing methods on various intrinsic and extrinsic tasks and provide an advantage when used with ELMo. 
We also propose \methodside{}, an effective framework for incorporating diverse semantic knowledge for further enhancing learned word representations. 
We make the source code of both models
available to encourage reproducible research.

\end{abstract}

\section{Introduction}
\label{sec:intro}
 
\begin{figure*}[t]
	\centering
	\includegraphics[scale=0.7]{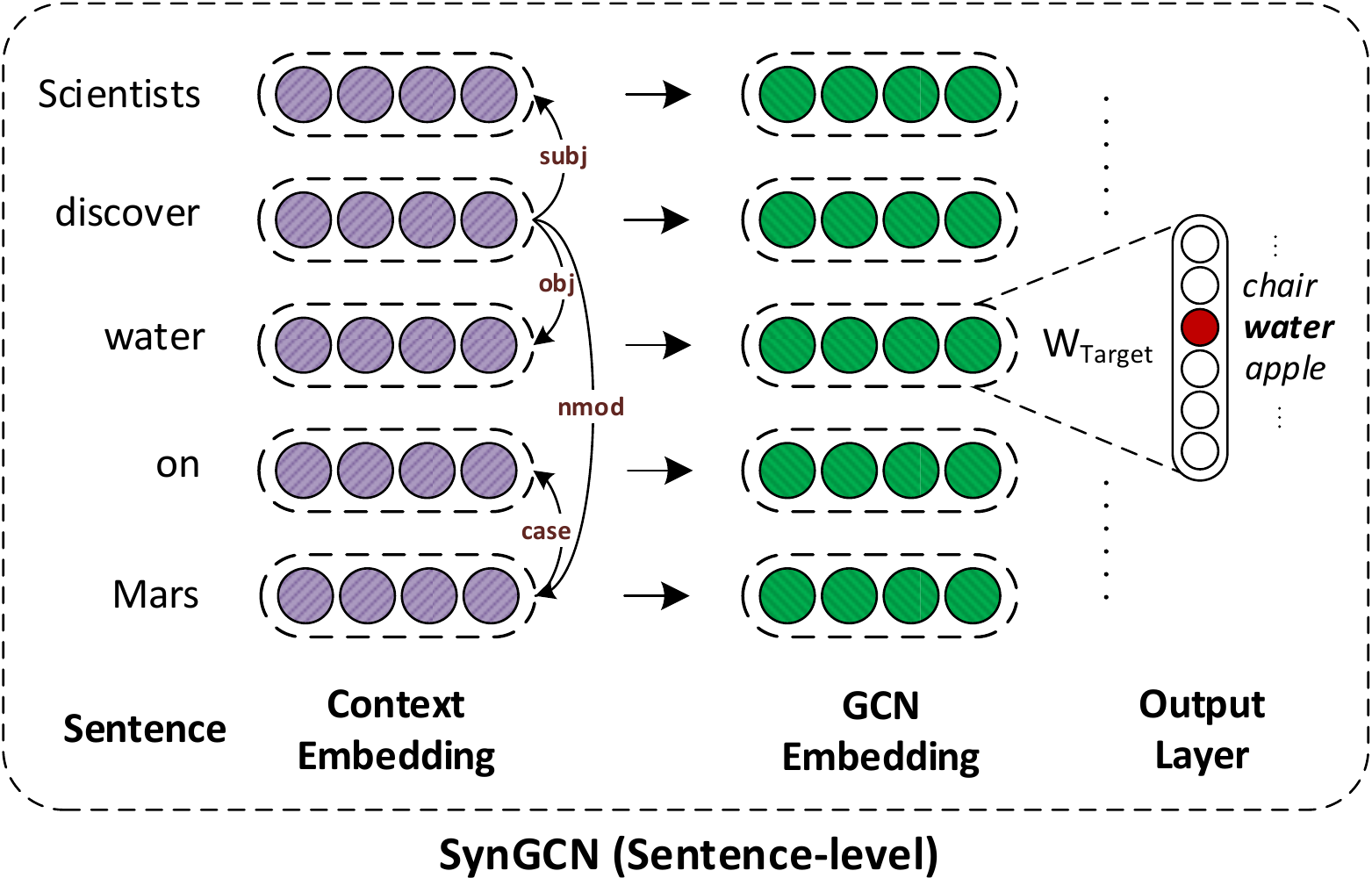}
	\caption{\label{fig:wordgcn_overview} Overview of \method{}: 
	\method{} employs Graph Convolution Network for utilizing dependency context for learning word embeddings. For each word in vocabulary, the model learns its representation by aiming to predict each word based on its dependency context encoded using GCNs. Please refer Section \ref{sec:wordgcn_details} for more details. 
}  
\end{figure*}

Representing words as real-valued vectors is an effective and widely adopted technique in NLP. Such representations capture 
properties of words based on their usage and allow them to generalize across tasks. Meaningful word embeddings have been shown to improve performance on several relevant tasks, such as named entity recognition (NER) \cite{app_ner}, parsing \cite{parsing_app}, and part-of-speech (POS) tagging \cite{pos_tagging}. Using word embeddings for initializing Deep Neural Networks has also been found to be quite useful \cite{Collobert2011,googleMT,lisa_paper}. 

Most popular methods for learning word embeddings are based on the distributional hypothesis, which utilizes the co-occurrence statistics from \emph{sequential} context of words for learning word representations \cite{Mikolov2013a,Pennington2014}. More recently, this approach has been extended to include syntactic contexts \cite{deps_paper} derived from dependency parse of text. Higher order dependencies have also been exploited by \citet{ext_paper,Li2018a}. Syntax-based embeddings encode functional similarity (in-place substitutable words) rather than topical similarity (topically related words) which provides an advantage on specific tasks like question classification \cite{ext_paper}.
However, current approaches incorporate syntactic context by concatenating words with their dependency relations. 
 For instance, in Figure \ref{fig:wordgcn_overview} \textit{scientists\_subj}, \textit{water\_obj}, and \textit{mars\_nmod} needs to be included as a part of vocabulary for utilizing the dependency context of \textit{discover}. 
 This severely expands the vocabulary, thus limiting the
scalability of models on large corpora. For instance, in \citet{deps_paper} and \citet{ext_paper}, the context vocabulary explodes to around 1.3 million for learning embeddings of 220k words.

Incorporating relevant signals from semantic knowledge sources such as WordNet \cite{wordnet_paper}, FrameNet \cite{framenet}, and Paraphrase Database (PPDB) \cite{ppdb_paper} has been shown to improve the quality of word embeddings. Recent works utilize these by incorporating them in a neural language modeling objective function 
\cite{Mo2014,japanese2018}, or as a post-processing step \cite{Faruqui2014,counter_paper}. Although existing approaches improve the quality of word embeddings, they require explicit modification for handling different types of semantic information. 

 
Recently proposed Graph Convolutional Networks (GCN) \cite{Defferrard2016,Kipf2016} have been found to be useful for encoding structural information in graphs. Even though GCNs have been successfully employed for several NLP tasks such as machine translation \cite{gcn_mt}, semantic role labeling \cite{gcn_srl}, document dating \cite{neuraldater} and text classification \cite{gcn_text_class}, they have so far not been used for learning word embeddings, especially leveraging cues such as syntactic and semantic information. GCNs provide flexibility to represent diverse syntactic and semantic relationships between words all within one framework, without requiring relation-specific special handling as in previous methods.  
Recognizing these benefits, we make the following contributions in this paper. 

 
\noindent
\begin{enumerate}[itemsep=2pt,parsep=0pt,partopsep=0pt,leftmargin=*,topsep=0.1pt]
	\item We propose \method{}, a Graph Convolution based method for learning word embeddings. Unlike previous methods, \method{} 
		utilizes syntactic context for learning word representations without increasing vocabulary size. 
	\item We also present \methodside{}, a framework for incorporating diverse semantic knowledge (e.g., synonymy, antonymy, hyponymy, etc.) in learned word embeddings, without requiring relation-specific special handling as in previous methods.
	\item Through experiments on multiple intrinsic and extrinsic tasks, we demonstrate that our proposed methods obtain substantial improvement over state-of-the-art approaches,  
	and also yield an advantage when used in conjunction with methods such as ELMo \cite{elmo_paper}.
\end{enumerate}

\noindent The source code of both the methods has been made available at \url{http://github.com/malllabiisc/WordGCN}.

\section{Related Work}
\label{sec:related}

\textbf{Word Embeddings:} Recently, there has been much interest in learning meaningful word representations such as neural language modeling \cite{Bengio2003} based continuous-bag-of-words (CBOW) and skip-gram (SG) models \cite{Mikolov2013a}. 
This is further extended by \citet{Pennington2014} which learns embeddings by factorizing word co-occurrence matrix to leverage global statistical information.  
Other formulations for learning word embeddings include multi-task learning \cite{Collobert2011} and ranking frameworks \cite{Ji2015}.



\textbf{Syntax-based Embeddings:} Dependency parse context based word embeddings is first introduced by \citet{deps_paper}. They allow encoding syntactic relationships between words and show improvements on tasks where functional similarity is more relevant than topical similarity. The inclusion of syntactic context is further enhanced through second-order \cite{ext_paper} and multi-order \cite{Li2018a} dependencies. However, in all these existing approaches, the word vocabulary is severely expanded for incorporating syntactic relationships. 


\textbf{Incorporating Semantic Knowledge Sources:}
Semantic relationships such as \textit{synonymy}, \textit{antonymy}, \textit{hypernymy}, etc. from several semantic sources have been utilized for improving the quality of word representations. Existing methods either exploit them jointly \cite{rc_net_joint2014,joint_app2015,japanese2018} or as a post-processing step \cite{Faruqui2014,counter_paper}. 
	\method{} falls under the latter category and is more effective at incorporating semantic constraints (Section \ref{sec:diverse_side_results} and \ref{sec:rfgcn_results}). 

\textbf{Graph Convolutional Networks:} 
In this paper, we use the first-order formulation of GCNs via a layer-wise propagation rule as proposed by \cite{Kipf2016}. Recently, some variants of GCNs have also been proposed \cite{lovasz_paper,confgcn}. A detailed description of GCNs and their applications can be found in \citet{Bronstein2017}. In NLP, GCNs have been utilized for semantic role labeling \cite{gcn_srl}, machine translation \cite{gcn_mt}, and relation extraction \cite{reside}. Recently, \citet{gcn_text_class} use GCNs for text classification by jointly embedding words and documents. However, their learned embeddings are task specific whereas in our work we aim to learn task agnostic word representations.




\section{Background: Graph Convolutional Networks}
\label{sec:background}

In this section, we will provide a brief overview of Graph Convolutional Networks (GCNs) \cite{Defferrard2016,Kipf2016} and its extension to directed labeled graphs.

\subsection{GCN on Directed Labeled Graphs}
\label{sec:gcn_directed}

Let $\mathcal{G}=(\mathcal{V},\mathcal{E}, \mathcal{X})$ be a directed graph where $\mathcal{V}$ is the set of nodes $(|\mathcal{V}|=n)$, $\mathcal{E}$ indicates the edge set, and 
$\mathcal{X} \in \mathbb{R}^{n\times d}$ denotes the $d$-dimensional \emph{input} node features. An edge from node $u$ to $v$ with label $l_{uv}$ is denoted by $(u,v,l_{uv})$.
%
 As the information need not always propagate only along the direction of the edge, following \citet{gcn_srl}, we include inverse edges $(v,u,l^{-1}_{uv})$ in $\mathcal{E}$. 
 Embedding $h_{v}^{k+1} \in \mathbb{R}^{d}$ of a node $v$ after $k$-GCN layers is given as follows.
 \begingroup\makeatletter\def\f@size{10}\check@mathfonts	
\begin{eqnarray*}
	h_{v}^{k+1} &=& f \left(\sum_{u \in \mathcal{N_{+}}(v)}\left(W^{k}_{l_{uv}}  h_{u}^{k} + b^{k}_{l_{uv}}\right) \right) \\
\end{eqnarray*}
\endgroup
Here, $W^{k}_{l_{uv}} \in \mathbb{R}^{d\times d}$ and $b_{l_{uv}} \in \mathbb{R}^d$ are label specific model parameters, $\mathcal{N_+}(v) = \mathcal{N}(v) \cup \{v\}$ is the set of immediate neighbors of $v$ (including $v$ itself), and $h^k_u \in \mathbb{R}^{d}$ is hidden representation of node $u$ after $k-1$ layers. \\

\noindent \textbf{Edge Label Gating Mechanism}: In real-world graphs, some of the edges might be erroneous or irrelevant for the downstream task. This is predominant in automatically constructed graphs like dependency parse of text.  
To address this issue, we employ edge-wise gating \cite{gcn_srl} in GCNs. For each node $v$, we calculate a relevance score $g^k_{l_{uv}} \in \mathbb{R}$ for all the edges in which $v$ participates. The score is computed independently for each layer as shown below.
\begingroup\makeatletter\def\f@size{10}\check@mathfonts	
\begin{equation*}
\label{eqn:edge_gating}
g^{k}_{l_{uv}} = \sigma \left( \hat{W}^{k}_{l_{uv}} h^{k}_u  + \hat{b}^{k}_{l_{uv}} \right)
\end{equation*}

Here, $\hat{W}^{k}_{l_{uv}} \in \mathbb{R}^{1 \times d}$ and $\hat{b}^{k}_{l_{uv}} \in \mathbb{R}$ are trainable parameters and $\sigma (\cdot)$ is the sigmoid function. The updated GCN propagation rule for the $k^{th}$ layer can be written as shown below.
\begingroup\makeatletter\def\f@size{10}\check@mathfonts	
\begin{equation}
\label{eqn:gcn_directed_gated}
h^{k+1}_{v} = f  \left(\sum_{u \in \mathcal{N_{+}}(v)} g^{k}_{l_{uv}} \times \left({W}^{k}_{l_{uv}} h^{k}_{u} + b^{k}_{l_{uv}}\right)\right)
\end{equation}
\endgroup

\section{Methods Overview}
\label{sec:overview}

%


The task of learning word representations in an unsupervised setting can be formulated as follows: Given a text corpus,
the aim is to learn a $d$-dimensional embedding for each word in the vocabulary. 
Most of the distributional hypothesis based approaches
only utilize sequential context for each word in the corpus. However, this becomes suboptimal when the relevant context words lie beyond the window size. For instance in \reffig{fig:wordgcn_overview}, a relevant context word \textit{discover} for \textit{Mars} is missed if the chosen window size is less than $3$. 
On the contrary, a large window size might allow irrelevant words to influence word embeddings negatively. 

Using dependency based context helps to alleviate this problem. However, all existing syntactic context based methods  \cite{deps_paper,ext_paper,Li2018a} severely expand vocabulary size (as discussed in Section \ref{sec:intro}) which  limits their scalability to a large corpus. To eliminate this drawback, we propose \method{} which employs Graph Convolution Networks 
to better encode syntactic information in embeddings. We prefer GCNs over other graph encoding architectures such as Tree LSTM \cite{tree_lstm} as GCNs do not restrict graphs to be trees and have been found to be more effective at capturing global information \cite{gcn_re_stanford}. Moreover, they give substantial speedup as they do not involve recursive operations which are difficult to  parallelize. The overall architecture is shown in Figure \ref{fig:wordgcn_overview}, for more details refer to Section \ref{sec:wordgcn_details}. 

Enriching word embeddings with semantic knowledge helps to improve their quality for several NLP tasks. 
Existing approaches are either incapable of utilizing these diverse relations or need to be explicitly modeled for exploiting them. 
In this paper, we propose \methodside{} which automatically learns to utilize multiple semantic constraints by modeling them as different edge types. 
It can be used as a post-processing method similar to \citet{Faruqui2014,counter_paper}. We describe it in more detail in Section \ref{sec:rfgcn_details}.


\section{\method{}}
\label{sec:wordgcn_details}
In this section, we provide a detailed description of our proposed method, \method{}. 
Following \citet{Mikolov2013b,deps_paper,ext_paper}, we separately define target and context embeddings for each word in the vocabulary as parameters in the model.
For a given sentence $s = (w_1, w_2, \ldots, w_n)$, we first extract its dependency parse graph $\mathcal{G}_s= (\mathcal{V}_s, \mathcal{E}_s)$ using Stanford CoreNLP parser \cite{stanford_corenlp}. Here, $\mathcal{V}_s = \{w_1, w_2, \dots, w_n\}$ and $\mathcal{E}_s$ denotes the labeled directed dependency edges of the form $(w_i, w_j, l_{ij})$, where $l_{ij}$ is the dependency relation of $w_i$ to $w_j$. 

Similar to \citet{Mikolov2013b}'s continuous-bag-of-words (CBOW) model, which defines the context of a word $w_i$ as 
$\m{C}_{w_i} = \{w_{i+j} : -c \leq j \leq c, j \neq 0\}$ for a window of size $c$, we define the context as its neighbors in $\m{G}_s$, i.e., $\m{C}_{w_i} = \m{N}(w_i)$. Now, unlike CBOW which takes the sum of the context embedding of words in $\m{C}_{w_i}$ to predict $w_i$, we apply directed Graph Convolution Network (as defined in Section \ref{sec:background}) on $\m{G}_s$ with context embeddings of words in $s$ as input features. Thus, for each word $w_i$ in $s$, we obtain a representation $h^{k+1}_i$ after $k$-layers of GCN using \refeqn{eqn:gcn_directed_gated} which we reproduce below for ease of readability (with one exception as described below).
\begin{equation*}
h^{k+1}_{i} = f  \left(\sum_{j \in \mathcal{N}(i)} g^{k}_{l_{ij}} \times \left({W}^{k}_{l_{ij}} h^{k}_{j} + b^{k}_{l_{ij}}\right)\right)
\end{equation*}

%
Please note that unlike in \refeqn{eqn:gcn_directed_gated}, we use 
	$\mathcal{N}({i})$ instead of $\mathcal{N_{+}}({i})$ in \method{}, i.e., we do not include self-loops in $\mathcal{G}_s$.
This helps to avoid overfitting to the initial embeddings, which is undesirable in the case of \method{} as it uses random initialization. We note that similar strategy has been followed by \citet{Mikolov2013b}. 
Furthermore, to handle erroneous edges in automatically constructed dependency parse graph, we perform edge-wise gating (\refsec{sec:gcn_directed}) to give importance to relevant edges and suppress the noisy ones. The embeddings obtained are then used to calculate the loss as described in \refsec{sec:loss}.

 \method{} utilizes \textit{syntactic context} to learn more meaningful word representations. We validate this in Section \ref{sec:intrinsic_results}. Note that, the word vocabulary remains unchanged during the entire learning process, this makes \method{} more scalable compared to the existing approaches.
 
 Note that, \method{} is a generalization of CBOW model, as shown below.


\begin{theorem}
\label{th:cbow_generalization}
\method{} is a generalization of Continuous-bag-of-words (CBOW) model.
\end{theorem}
\begin{proof}
The reduction can be obtained as follows. For a given sentence $s$, take the neighborhood of each word $w_i$ in $\m{G}_s$ as it sequential context, i.e., $\m{N}(w_i) = \{w_{i+j} : -c \leq j \leq c, j \neq 0 \} \ \forall w_i \in s$. Now, if the number of GCN layers are restricted to $1$ and the activation function is taken as identity $(f(x) =x)$, then \refeqn{eqn:gcn_directed_gated} reduces to
\[
h_{i} = \sum_{-c \leq j \leq c, j \neq 0} \left(g_{l_{ij}} \times \left({W}_{l_{ij}} h_{j} + b^{k}_{l_{ij}}\right)  \right). 
\]
Finally, $W^{k}_{l_{ij}}$ and $b^{k}_{l_{ij}}$ can be fixed to an identity matrix $(\mathbb{\bold{I}})$ and a zero vector $(\bold{0})$, respectively, and edge-wise gating $(g_{l_{ij}})$ can be set to $1$. This gives 
\[
h_{i} = \sum_{-c \leq j \leq c, j \neq 0} \left( \mathbb{\bold{I}} \cdot h_{j} + \bold{0} \right)= \sum_{-c \leq j \leq c, j \neq 0} h_{j} ,
\]
which is the hidden layer equation of CBOW model. 
\end{proof}

\section{\methodside{}}
\label{sec:rfgcn_details} 
In this section, we propose another Graph Convolution based framework, \methodside{}, for incorporating semantic knowledge in pre-trained word embeddings. 
Most of the existing approaches 
like \citet{Faruqui2014,counter_paper} are restricted to handling symmetric relations like \textit{synonymy} and \textit{antonymy}. On the other hand, although recently proposed \cite{japanese2018} is capable of handling  asymmetric information, it still requires manually defined relation strength function which can be labor intensive and suboptimal. 

\methodside{} is capable of incorporating both symmetric as well as asymmetric information jointly. Unlike \method{}, \methodside{} operates on a corpus-level directed labeled graph with words as nodes and edges representing semantic relationship among them from different sources. For instance, in Figure \ref{fig:rfgcn_model}, semantic relations such as \textit{hyponymy}, \textit{hypernymy} and \textit{synonymy} are represented together in a single graph. Symmetric information is handled by including a directed edge in both directions. Given the corpus level graph $\m{G}$, the training procedure is similar to that of \method{}, i.e.,  predict the word $w$ based on its neighbors in $\m{G}$. Inspired by \citet{Faruqui2014}, we preserve the semantics encoded in pre-trained embeddings by initializing both target and context embeddings with given word representations and keeping target embeddings fixed during training. \methodside{} uses \refeqn{eqn:gcn_directed_gated} to update node embeddings. Please note that in this case $\mathcal{N_{+}}(v)$ is used as the neighborhood definition to preserve the initial learned representation of the words.

\label{sec:details}
\begin{figure}[t]
	\centering
	\includegraphics[scale=0.6]{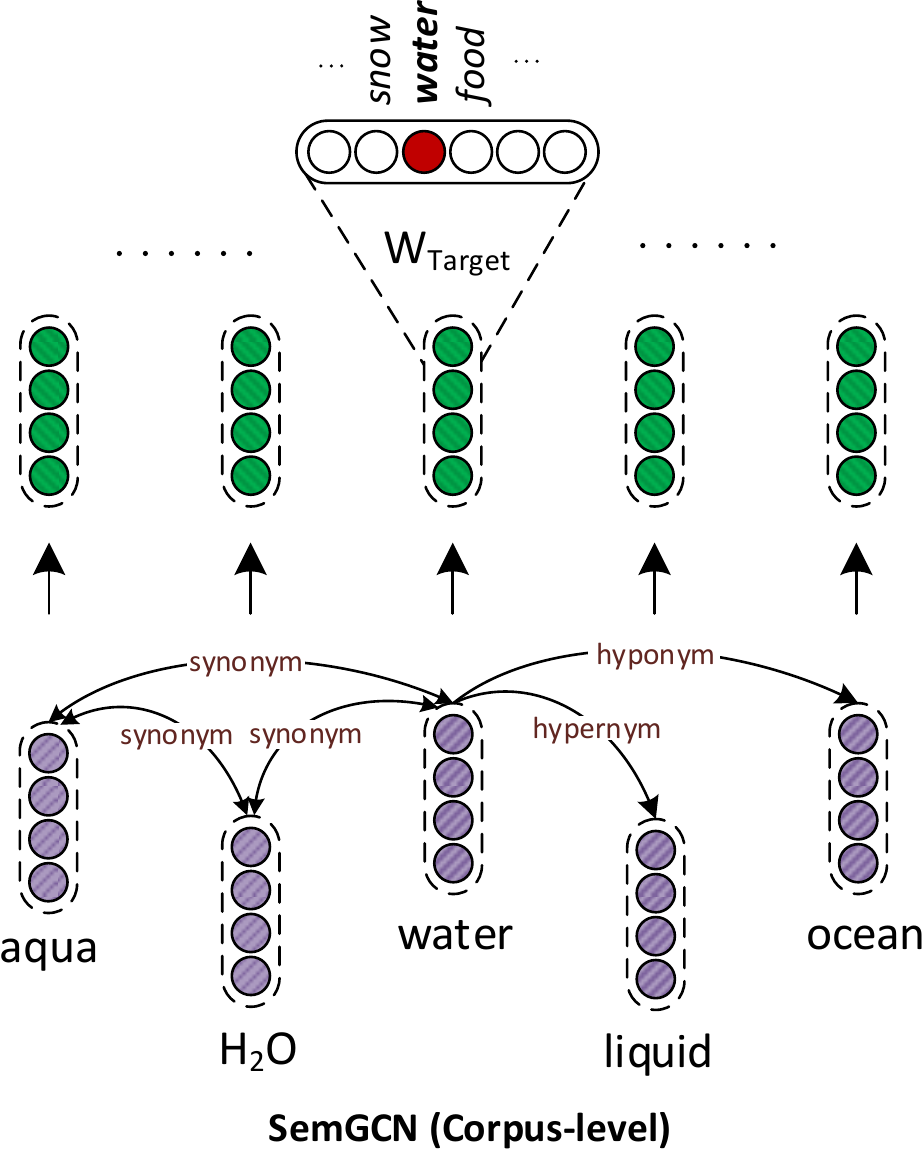}
	\caption{\label{fig:rfgcn_model}Overview of \methodside{}, our proposed Graph Convolution based framework for incorporating diverse semantic information in learned embeddings. Double-headed edges denote two edges in both directions. Please refer to Section \ref{sec:rfgcn_details} for more details.}  
\end{figure}

\section{Training Details} 
\label{sec:loss}
Given the GCN representation $(h_t)$ of a word $(w_t)$, the training objective of \method{} and \methodside{} is to predict the target word given its neighbors in the graph. Formally, for each method we maximize the following objective\footnote{We also experimented with joint \method{} and \methodside{} model but our preliminary experiments gave suboptimal performance as compared to the sequential model. This can be attributed to the fact that syntactic information is orders of magnitude greater than the semantic information available. Hence, the semantic constraints are not effectively utilized. We leave the analysis of the joint model as a future work.}.
 \begingroup\makeatletter\def\f@size{10}\check@mathfonts	
\[
	E = \sum_{t=1}^{|V|} \log P(w_t | w^{t}_{1}, w^{t}_{2} \dots w^{t}_{N_t})
\] 
\endgroup
where, $w_t$ is the target word and $w^{t}_{1}, w^{t}_{2} \dots w^{t}_{N_t}$ are its neighbors in the graph. 
The probability $P(w_t | w^{t}_{1}, w^{t}_{2} \dots w^{t}_{N_t})$ is calculated using the softmax function, defined as
 \begingroup\makeatletter\def\f@size{10}\check@mathfonts	
\[	
P(w_t | w^{t}_{1}, w^{t}_{2} \dots w^{t}_{N_t}) = \frac{\exp(v^T_{w_t} h_t)}{\sum_{i=1}^{|V|} \exp(v^T_{w_{i} }h_t)}.
\]
\endgroup
Hence, $E$ reduces to 
 \begingroup\makeatletter\def\f@size{10}\check@mathfonts	
\begin{equation}
E = \sum_{t=1}^{|V|}  \left( v^T_{w_t} h_t - \log \sum_{i=1}^{|V|} \exp(v^T_{w_{i} }h_t ) \right),
\label{eqn:full_softmax}
\end{equation}
\endgroup
where, $h_t$ is the GCN representation of the target word $w_t$ and $v_{w_t}$ is its target embedding.

The second term in \refeqn{eqn:full_softmax} is computationally expensive as the summation needs to be taken over the entire vocabulary. This can be overcome using several approximations like noise-contrastive estimation \cite{gutmann10a} and hierarchical softmax \cite{hierarchical_softmax1}. In our methods, we use negative sampling as used by \citet{Mikolov2013b}.

\section{Experimental Setup}
\label{sec:experiments}

\subsection{Dataset and Training}
\label{sec:dataset}
In our experiments, we use Wikipedia\footnote{https://dumps.wikimedia.org/enwiki/20180301/} corpus for training the models.  After discarding too long and too short sentences, we get an average sentence length of nearly $20$ words. The corpus consists of 57 million sentences with 1.1 billion tokens and 1 billion syntactic dependencies. 

\subsection{Baselines}
\label{sec:baselines}

For evaluating \method{} (Section \ref{sec:wordgcn_details}), we compare against the following baselines:
\begin{itemize}[itemsep=1pt,parsep=0pt,partopsep=0pt,leftmargin=10pt,topsep=1pt]
	\item \textbf{Word2vec} is continuous-bag-of-words model originally proposed by \citet{Mikolov2013b}.
	\item \textbf{GloVe} \cite{Pennington2014}, a log-bilinear regression model which leverages global co-occurrence statistics of corpus.
	\item \textbf{Deps} \cite{deps_paper} is a modification of skip-gram model which uses dependency context in place of sequential context. 
	\item \textbf{EXT} \cite{ext_paper} is an extension of Deps which utilizes second-order dependency context features. 	 
\end{itemize}

\noindent \methodside{} (Section \ref{sec:rfgcn_details}) model is evaluated against the following methods:
\begin{itemize}[itemsep=1pt,parsep=0pt,partopsep=0pt,leftmargin=10pt,topsep=2pt]

	\item \textbf{Retro-fit} \cite{Faruqui2014} is a post-processing procedure which uses similarity constraints from semantic knowledge sources. 
	\item \textbf{Counter-fit} \cite{counter_paper}, a method for injecting both antonym and synonym constraints into word embeddings.
	\item \textbf{JointReps} \cite{japanese2018}, a joint word representation learning method which simultaneously utilizes the corpus and KB.
\end{itemize}

\subsection{Evaluation method}
\label{sec:evaluation_tasks}
To evaluate the effectiveness of our proposed methods,  we compare them against the baselines on the following intrinsic and extrinsic tasks\footnote{Details of hyperparameters are in supplementary.}: 

\begin{itemize}[itemsep=2pt,parsep=0pt,partopsep=0pt,leftmargin=*,topsep=2pt]
\item \textbf{Intrinsic Tasks:} \\	
\noindent \textbf{Word Similarity} is the task of evaluating closeness between semantically similar words.  Following \citet{ext_paper,Pennington2014}, we evaluate on 
Simlex-999 \cite{simlex_dataset}, 
WS353
\cite{ws353_dataset}, 
and RW \cite{rw_dataset} datasets.

\noindent \textbf{Concept Categorization} involves grouping nominal concepts into natural categories. For instance, \textit{tiger} and \textit{elephant} should belong to \textit{mammal} class. In our experiments, we evalute on AP \cite{ap_dataset}, Battig \cite{battig_dataset}, BLESS \cite{bless_dataset}, ESSLI \cite{essli_dataset} datasets.

\noindent \textbf{Word Analogy} task is to predict word $b_{2}$, given three words $a_{1}$, $a_{2}$, and $b_{1}$, such that the relation $b_{1}:b_{2}$ is same as the relation $a_{1}:a_{2}$. We compare methods on MSR \cite{msr_dataset} and SemEval-2012 \cite{semeval_dataset}.



\begin{table*}[tbh]
	\centering
	\begin{small}
		\begin{tabular}{lcccc|cccc|cc}
			\toprule
			& \multicolumn{4}{c}{Word Similarity} & \multicolumn{4}{c}{Concept Categorization} & \multicolumn{2}{c}{Word Analogy} \\ 
			\cmidrule(r){2-5} \cmidrule(r){6-9} \cmidrule(r){10-11} 
			Method        & WS353S        & WS353R        & SimLex999     & RW            & AP            & Battig        & BLESS         & ESSLI     & SemEval2012   & MSR           \\
			\midrule
			Word2vec      & 71.4          & 52.6          & 38.0          & 30.0          & 63.2          & 43.3          & 77.8          & 63.0          & 18.9          & 44.0          \\
			GloVe         & 69.2          & \bf{53.4}     & 36.7          & 29.6          & 58.0          & 41.3          & 80.0          & 59.3          & 18.7          & 45.8          \\
			Deps          & 65.7          & 36.2          & 39.6          & 33.0          & 61.8          & 41.7          & 65.9          & 55.6          & 22.9          & 40.3          \\
			EXT           & 69.6          & 44.9          & 43.2          & 18.6          & 52.6          & 35.0          & 65.2          & 66.7          & 21.8          & 18.8          \\
			\midrule
			\method{}  & \bf{73.2}     & 45.7          & \bf{45.5}     & \bf{33.7}     & \bf{69.3}     & \bf{45.2}     & \bf{85.2}     & \bf{70.4}     & \bf{23.4}     & \bf{52.8}     \\
			\bottomrule
		\end{tabular}
	\end{small}
	\caption{\label{tbl:intrinsic} \textbf{\method{} Intrinsic Evaluation:} Performance on word similarity (Spearman correlation), concept categorization (cluster purity), and word analogy (Spearman correlation). Overall, \method{} outperforms other existing approaches in $9$ out of $10$ settings. Please refer to \refsec{sec:intrinsic_results} for more details. 
	}
\end{table*}

\item \textbf{Extrinsic Tasks:} \\
\noindent \textbf{Named Entity Recognition (NER)} is the task of locating and classifying entity mentions into categories like \textit{person}, \textit{organization} etc. We use \citet{e2e_coref_model}'s model on CoNLL-2003 dataset \cite{conll03_dataset} for evaluation.

\noindent \textbf{Question Answering} in Stanford Question Answering Dataset (\textbf{SQuAD}) \cite{squad_dataset} involves identifying answer to a question as a segment of text from a given passage. Following \citet{elmo_paper}, we evaluate using \citet{docqa_model}'s model. 

\noindent \textbf{Part-of-speech (POS) tagging} aims at associating with each word, a unique tag describing its syntactic role. For evaluating word embeddings, we use \citet{e2e_coref_model}'s model on Penn Treebank POS dataset \cite{penn_pos_dataset}.

\noindent \textbf{Co-reference Resolution (Coref)} involves identifying all expressions that refer to the same entity in the text. To inspect the effect of embeddings, we use \citet{e2e_coref_model}'s model on CoNLL-2012 shared task dataset  \cite{coref_conll12_dataset}.
\end{itemize}

\section{Results}
\label{sec:results}

In this section, we attempt to answer the following questions.
\begin{itemize}[itemsep=2pt,topsep=2pt,parsep=1pt,partopsep=0pt]
	\item[Q1.] Does \method{} learn better word embeddings than existing approaches? (Section \ref{sec:intrinsic_results})
	\item[Q2.] Does \methodside{} effectively handle diverse semantic information as compared to other methods? (Section \ref{sec:diverse_side_results})
	\item[Q3.] How does \methodside{} perform compared to other methods when provided with the same semantic constraints? (Section \ref{sec:rfgcn_results})
	\item[Q4.] Does dependency context based embedding encode complementary information compared to ELMo? (Section \ref{sec:elmo_comp})
\end{itemize}

\subsection{\method{} Evaluation}
\label{sec:intrinsic_results}

The evaluation results on intrinsic tasks -- word similarity, concept categorization, and analogy --  are summarized in \reftbl{tbl:intrinsic}. We report Spearman correlation for word similarity and analogy tasks and cluster purity for concept categorization task. Overall, we find that  \method{}, our proposed method, outperforms all the existing word embedding approaches in 9 out of 10 settings. The inferior performance of \method{} and other dependency context based methods on WS353R dataset compared to sequential context based methods is consistent with the observation reported in \citet{deps_paper,ext_paper}. This is because the syntactic context based embeddings capture functional similarity rather than topical similarity (as discussed in Section \ref{sec:intro}). On average, we obtain around $1.5$\%,  $5.7$\%  and $7.5$\% absolute increase in performance on word similarity, concept categorization and analogy tasks compared to the best performing baseline. The results demonstrate that the learned embeddings from \method{} more effectively capture semantic and syntactic properties of words.

We also evaluate the performance of different word embedding approaches on the downstream tasks as defined in \refsec{sec:evaluation_tasks}. 
The experimental results are summarized in Table \ref{tbl:extrinsic}. Overall, we find that \method{} either outperforms or performs comparably to other methods on all four tasks. Compared to the sequential context based methods, dependency based methods perform superior at question answering task as they effectively encode syntactic information. This is consistent with the observation of \citet{elmo_paper}. 

\begin{table}[t]
	\centering
	\resizebox{\columnwidth}{!}{
		\begin{tabular}{lcccc}
			\toprule
			Method          & POS                   & SQuAD                 & NER                   & Coref \\ 
			\midrule
			Word2vec        & 95.0$\pm$0.1          & 78.5$\pm$0.3          & 89.0$\pm$0.2          & 65.1$\pm$0.3 \\ 
			GloVe           & 94.6$\pm$0.1          & 78.2$\pm$0.2          & 89.1$\pm$0.1          & 64.9$\pm$0.2 \\ 
			Deps            & 95.0$\pm$0.1          & 77.8$\pm$0.3          & 88.6$\pm$0.3          & 64.8$\pm$0.1 \\ 
			EXT             & 94.9$\pm$0.2          & \bf{79.6$\pm$0.1}     & 88.0$\pm$0.1          & 64.8$\pm$0.1 \\ 
			\midrule
			\method{}      	& \bf{95.4$\pm$0.1}     & \bf{79.6$\pm$0.2}     & \bf{89.5$\pm$0.1}     & \bf{65.8$\pm$0.1 }\\ 
			\bottomrule
		\end{tabular}	
	}
	\caption{\label{tbl:extrinsic} \textbf{\method{} Extrinsic Evaluation:} Comparison on parts-of-speech tagging (POS), question answering (SQuAD), named entity recognition (NER), and co-reference resolution (Coref). \method{} performs comparable or outperforms all existing approaches on all tasks. Refer \refsec{sec:intrinsic_results} for details.}
\end{table}

\begin{table}[t]
	\centering
	\resizebox{\columnwidth}{!}{
		\begin{tabular}{lcccc}
			\toprule
			Method          & POS                   & SQuAD                 & NER                   & Coref \\ 
			\midrule
			X = \method{}      	& 95.4$\pm$0.1    		& 79.6$\pm$0.2   		& \textbf{89.5$\pm$0.1} & 65.8$\pm$0.1 \\ 
			Retro-fit (X,1)      & 94.8$\pm$0.1          & 79.6$\pm$0.1          & 88.8$\pm$0.1          & 66.0$\pm$0.2 \\ 
			Counter-fit (X,2)    & 94.7$\pm$0.1          & 79.8$\pm$0.1          & 88.3$\pm$0.3          & 65.7$\pm$0.3 \\ 
			JointReps (X,4)      & 95.4$\pm$0.1          & 79.4$\pm$0.3          & 89.1$\pm$0.3          & 65.6$\pm$0.1 \\ 
			\midrule
			\methodside{} (X,4)  & \textbf{95.5$\pm$0.1} & \bf{80.4$\pm$0.1}     & \textbf{89.5$\pm$0.1} & \textbf{66.1$\pm$0.1} \\ 
			\bottomrule
		\end{tabular}
	}
	\caption{\label{tbl:rfgcn_ext_full} \textbf{\methodside{} Extrinsic Evaluation:} Comparison of different methods for incorporating diverse semantic constraints in \method{} embeddings on all extrinsic tasks. Refer Section \ref{sec:diverse_side_results} for details.}
\end{table}

\begin{table*}[!t]
	\centering
	\resizebox{\textwidth}{!}{
		\begin{tabular}{l|ccc|ccc|ccc|ccc|ccc}
			\toprule
			\multicolumn{1}{c}{Init Embeddings (=X)}			& \multicolumn{3}{c}{Word2vec} 		& \multicolumn{3}{c}{GloVe} 		& \multicolumn{3}{c}{Deps} 			& \multicolumn{3}{c}{EXT} & \multicolumn{3}{c}{\method} 		\\ 
			\cmidrule(r){1-1} \cmidrule(r){2-4} \cmidrule(r){5-7} \cmidrule(r){8-10} \cmidrule(r){11-13} \cmidrule(r){14-16}
			Datasets		          & WS353      & AP         & MSR       & WS353       & AP        & MSR       & WS353      & AP        & MSR       & WS353      & AP         & MSR        & WS353      & AP        & MSR    \\
			\midrule
			Performance of X     & 63.0      & 63.2       & 44.0      & 58.0       & 60.4      & 45.8      & 55.6      & 64.2      & 40.3      & 59.3      & 53.5       & 18.8       & 61.7      & 69.3      & 52.8 \\
			\midrule
			Retro-fit (X,1)       & 63.4      & \bf{67.8}  & \bf{46.7} & 58.5       & 61.1      & \bf{47.2} & 54.8      & 64.7      & 41.0      & 61.6      & 55.1       & 40.5       & 61.2      & 67.1      & 51.4   \\
			Counter-fit (X,2)     & 60.3      & 62.9       & 31.4      & 53.7       & 62.5      & 29.6      & 46.9      & 60.4      & 33.4      & 52.0      & 54.4       & 35.8       & 55.2      & 66.4      & 31.7  \\
			JointReps (X,4)     & 60.9      & 61.1       & 28.5      & 59.2       & 55.5      & 37.6      & 54.8      & 58.7      & 38.0      & 58.8      & 54.8       & 20.6       & 60.9      & 68.2      & 24.9  \\
			\midrule
			\methodside{} (X,4)   & \bf{64.8} & \bf{67.8}  & 36.8      & \bf{63.3}  & \bf{63.2} & 44.1      & \bf{62.3} & \boxed{\bf{69.3}} & \bf{41.1} & \bf{62.9} & \bf{67.1}  & \bf{52.1}  & \boxed{\bf{65.3}} & \boxed{\bf{69.3}} & \boxed{\bf{54.4}} \\
			\bottomrule
		\end{tabular}
	}
	\caption{\label{tbl:rfgcn_intrinsic} \textbf{\methodside{} Intrinsic Evaluation:} Evaluation of different methods for incorporating diverse semantic constraints initialized using various pre-trained embeddings (X). M(X, R) denotes the fine-tuned embeddings using method M taking X as initialization embeddings. R denotes the type of semantic relations used as defined in Section \ref{sec:diverse_side_results}. \methodside{} outperforms other methods in $13$ our of $15$ settings. \methodside{} with \method{} gives the best performance across all tasks (highlighted using \boxed{\cdot}). Please refer Section \ref{sec:diverse_side_results} for details.}
\end{table*}

\subsection{Evaluation with Diverse Semantic Information}
\label{sec:diverse_side_results}

In this section, we compare \methodside{} against the methods listed in Section \ref{sec:baselines} for incorporating diverse semantic information in pre-trained embeddings. We use \textit{hypernym}, \textit{hyponym},  and \textit{antonym} relations from WordNet, and \textit{synonym} relations from PPDB as semantic information. 
For each method, we provide the semantic information that it can utilize, e.g., Retro-fit can only make use of \textit{synonym} relation\footnote{Experimental results controlling for semantic information are provided in \refsec{sec:rfgcn_results}.}. 
In our results, \textbf{M(X, R)} denotes the fine-tuned embeddings obtained using method M while taking X as initialization embeddings. R denotes the types of semantic information used as defined below.
\begin{itemize}[itemsep=0pt,parsep=0pt,partopsep=0pt,leftmargin=*,topsep=0pt]
	\item \textbf{R=1:} Only synonym information. 
	\item \textbf{R=2:} Synonym and antonym information. 
	\item \textbf{R=4:} Synonym, antonym, hypernym and hyponym information. 
\end{itemize}
For instance, Counter-fit (GloVe, 2) represents GloVe embeddings fine-tuned by Counter-fit using synonym and antonym information.

Similar to Section \ref{sec:intrinsic_results}, the evaluation is performed on the three intrinsic tasks. Due to space limitations, we report results on one representative dataset per task. 
The results are summarized in Table \ref{tbl:rfgcn_intrinsic}. We find that in $13$ out of $15$ settings, \methodside{} outperforms other methods. Overall, we observe that \methodside{}, when initialized with \method{}, gives the best performance on all the tasks (highlighted by \boxed{\cdot} in Table \ref{tbl:rfgcn_intrinsic}).

For comparing performance on the extrinsic  tasks, we first fine-tune \method{} embeddings using different methods for incorporating semantic information. The embeddings obtained by this process are then evaluated on extrinsic tasks, as in 
Section \ref{sec:intrinsic_results}. The results are shown in Table \ref{tbl:rfgcn_ext_full}. 
We observe that while the other methods do not always consistently give improvement over the baseline \methodsyn{}, \methodside{} is able to improve upon \methodsyn{} in all settings (better or comparable).
Overall, we observe that \method{} along with \methodside{} is the most suitable method for incorporating both syntactic and semantic information.

\begin{figure}[t]
	\centering
	\includegraphics[width=\columnwidth]{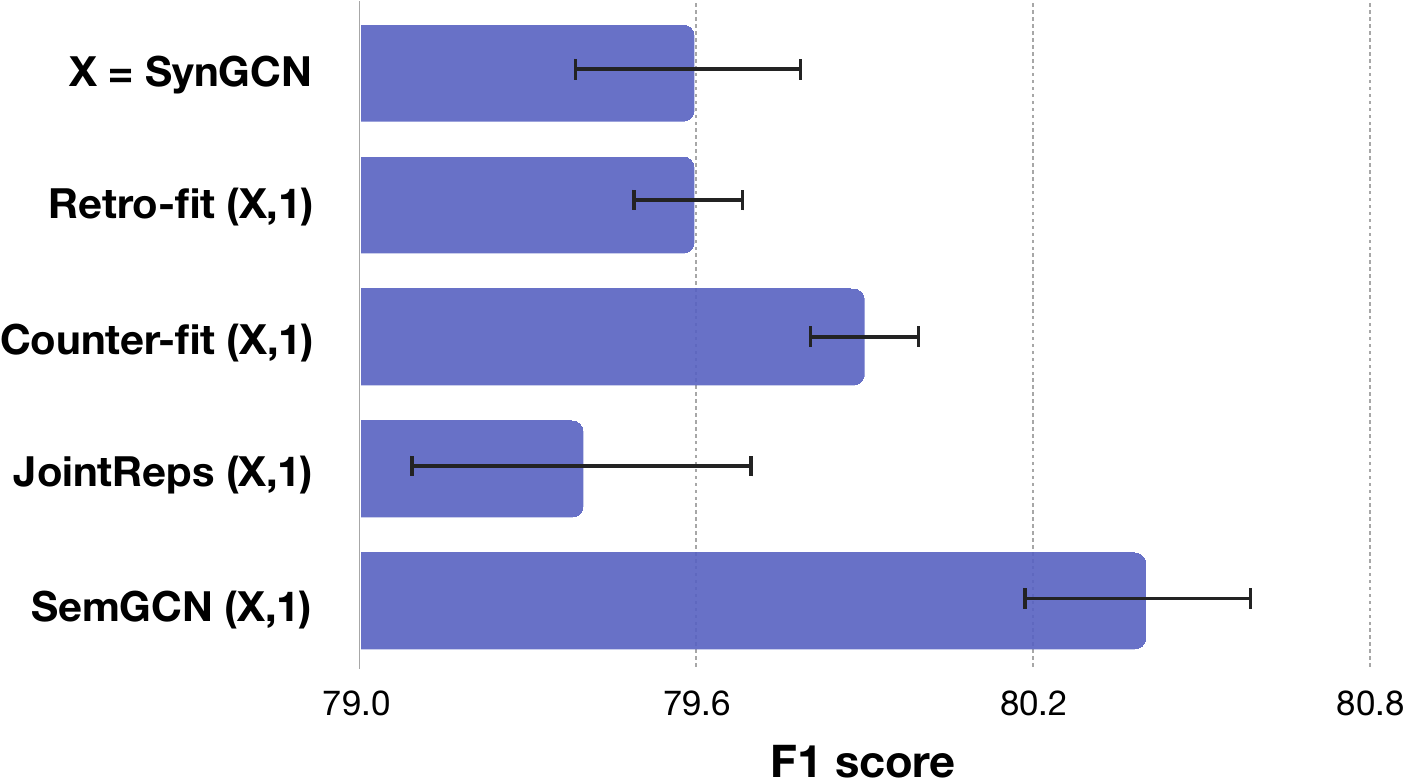}
	\caption{\label{fig:rfgcn_ppdb_ext}
		Comparison of different methods when provided with the same semantic information (synonym) for fine tuning \method{} embeddings. Results denote the F1-score on SQuAD dataset. \methodside{} gives considerable improvement in performance. Please refer Section \ref{sec:rfgcn_results} for details.
	} 
\vspace{-0.5mm}
\end{figure}

\subsection{Evaluation with Same Semantic Information}

\label{sec:rfgcn_results}
In this section, we compare \methodside{} against other baselines when provided with the same semantic information: synonyms from PPDB. Similar to Section \ref{sec:diverse_side_results}, we compare both on intrinsic and extrinsic tasks with different initializations. The evaluation results of fine-tuned \method{} embeddings by different methods on SQuAD are shown in the Figure \ref{fig:rfgcn_ppdb_ext}. The remaining results are included in the supplementary (Table S1 and S2). We observe that compared to other methods, \methodside{} is most effective at incorporating semantic constraints across all the initializations and outperforms others at both intrinsic and extrinsic tasks. 

\begin{table}[t]
	
	\centering
	\resizebox{\columnwidth}{!}{
		\begin{tabular}{lcccc}
			\toprule
			Method          & POS                   & SQuAD                 & NER                   & Coref \\ 
			\midrule
			ELMo (E)     										& 96.1$\pm$0.1    		& 81.8$\pm$0.2   		& 90.3$\pm$0.3 & 67.8$\pm$0.1 \\ 
		E+SemGCN(SynGCN, 4)  & \textbf{96.2$\pm$0.1}          & \textbf{82.4$\pm$0.1}          & \textbf{90.9$\pm$0.1} & \textbf{68.3$\pm$0.1} \\ 

			\bottomrule
		\end{tabular}
	}
	\caption{\label{tbl:elmo_comp} Comparison of ELMo with \method{} and \methodside{} embeddings on multiple extrinsic tasks. For each task, models use a linear combination of the provided embeddings whose weights are learned. Results show that our proposed methods encode complementary information which is not captured by ELMo. Please refer \refsec{sec:elmo_comp} for more details.}
\end{table}

\subsection{Comparison with ELMo}
\label{sec:elmo_comp}

Recently, ELMo \cite{elmo_paper} has been proposed which fine-tunes word embedding based on sentential context. In this section, we evaluate \method{} and \methodside{} when given along with pre-trained ELMo embeddings. The results are reported in Table \ref{tbl:elmo_comp}. The results show that dependency context based embeddings encode complementary information which is not captured by ELMo as it only relies on sequential context. Hence, our proposed methods serves as an effective combination with ELMo. 
\section{Conclusion}
\label{sec:conclusion}
In this paper, we have proposed \method{}, a graph convolution based approach which utilizes syntactic context for learning word representations. \method{} overcomes the problem of vocabulary explosion and outperforms state-of-the-art word embedding approaches on several intrinsic and extrinsic tasks. We also propose \methodside{}, a framework for jointly incorporating diverse semantic information in pre-trained word embeddings. The combination of \method{} and \methodside{} gives the best overall performance.
We make the source code of both models available to encourage reproducible research.


\section*{Acknowledgments}
We thank the anonymous reviewers for their constructive comments. This work is supported in part by the Ministry of Human Resource Development (Government of India) and Google PhD
 Fellowship.

\balance
\bibliography{acl2019}
\bibliographystyle{acl_natbib}

\end{document}